\documentclass[a4paper, 10 pt, conference]{ieeeconf}  

\IEEEoverridecommandlockouts    

\usepackage{graphics}
\usepackage{epsfig}
\usepackage{epstopdf}
\usepackage{lipsum}
\usepackage{amsmath}
\usepackage{algorithm}
\usepackage{algpseudocode}

\usepackage{url}
\title{\LARGE \bf Vehicle Trajectory Prediction in Crowded Highway Scenarios Using Bird Eye View Representations and CNNs}

\author{R. Izquierdo, A. Quintanar, I. Parra, D. Fern\'andez-Llorca, and M. A. Sotelo
\thanks{R. Izquierdo, A. Quintanar, I. Parra, D. Fern\'andez-Llorca and M. A. Sotelo are with the Computer Engineering Department, Universidad de Alcal\'a, Alcal\'a de Henares, Spain
        {\tt\small ruben.izquierdo@uah.es}}%
}

\begin{document}
\maketitle
\thispagestyle{empty}
\pagestyle{empty}

\begin{abstract}
This paper describes a novel approach to perform vehicle trajectory predictions employing graphic representations. The vehicles are represented using Gaussian distributions into a Bird Eye View. Then the U-net model is used to perform sequence to sequence predictions. This deep learning-based methodology has been trained using the HighD dataset, which contains vehicles' detection in a highway scenario from aerial imagery. The problem is faced as an image to image regression problem training the network to learn the underlying relations between the traffic participants. This approach generates an estimation of the future appearance of the input scene, not trajectories or numeric positions. An extra step is conducted to extract the positions from the predicted representation with subpixel resolution. Different network configurations have been tested, and prediction error up to three seconds ahead is in the order of the representation resolution. The model has been tested in highway scenarios with more than 30 vehicles simultaneously in two opposite traffic flow streams showing good qualitative and quantitative results.

\end{abstract}

\section{Introduction and Related Work}
Highways are among the most common driving scenarios in which autonomous vehicles are starting to develop their autonomous capabilities. Decision making is one of the most critical tasks of autonomous vehicles. Current decisions are based not only on what the vehicle is currently sensing but also on how it will evolve. Thus, the prediction of surrounding vehicles' trajectories becomes of utmost importance as underlying support for decision making. In the last years, this topic took relevance, and some datasets have been released to develop trajectory prediction models. These datasets are basically of two types regarding the recording point of view. The NGSIM datasets \cite{NGSIMI80}, \cite{NGSIMHW101}, and the HighD dataset \cite{highDdataset} provided vehicle trajectories recorded from an exterior point of view covering a static area over a highway. Other initiatives such as \cite{PKU_dataset}, \cite{lisa}, \cite{huang2018apolloscape}, \cite{waymo_open_dataset}, and \cite{prevention} uses mobile platforms to record surround vehicle trajectories. The problem with some of the in-vehicle dataset is that they are not specifically designed to develop trajectory predictions. The number of samples is insufficient, the data rate is insufficient, or their sensors do not provide the proper detection range. The off-vehicle datasets have an advantage that their data is not affected by occlusions, but they need to be adapted to be deployed on vehicles.

According to \cite{lefevre2014survey}, vehicle trajectory predictions can be classified into three levels: physical-based, maneuver-based, and intention-aware. The first type is based on mathematical models that fit the vehicle dynamics \cite{Houenou2013}. The second tries to predict the driver's intention and generates a trajectory corresponding to the predicted maneuver. The third type predicts trajectories modeling in some manner inter-dependencies between the traffic agents. Some works such as \cite{Ranjeet2010} or \cite{RubenITSC17} used Artificial Neural Networks to model the underlying behavior of a vehicle through their lateral positions. Some times trajectories are used to predict actions, and other actions are used to predict trajectories\cite{RubenITSC17}. Recently, more complex approaches that models vehicle interaction \cite{hu2018}, \cite{Deo2018}. \cite{Deo2018} uses the social pooling layer to create a connection between the predicted target and other traffic participants. The intrinsic state of the vehicles is commonly encoded using a Recurrent Neural Network(RNN).

In this paper, we propose a simple Bird Eye View representation and state-of-the-art CNN to make trajectory predictions in crowded highway scenarios. Interactions are not explicitly modeled, but they are considered included by itself.

\begin{figure}[t]
    \centering
    {
    \includegraphics[width=0.8\columnwidth]{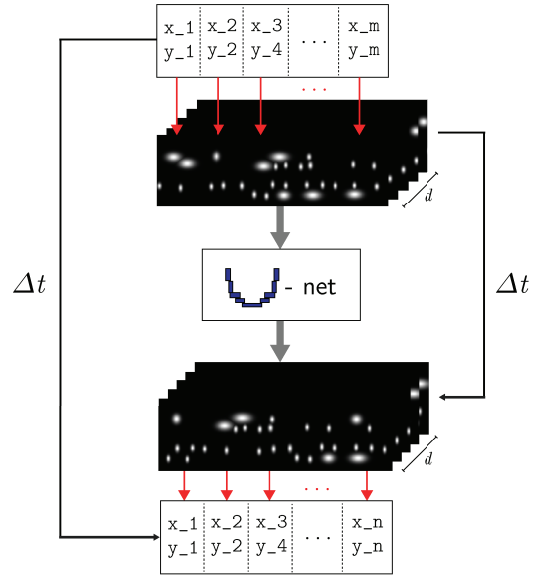}
    }
    \vspace{-2pt}
    \caption{System overview.}
    \vspace{-8pt}
    \label{fig:intro}
\end{figure}

The rest of the paper is organized as follows. Section II provides a thorough description of the network architecture. Section III describes how HighD dataset is transformed to be used as input and output of the proposed network architecture. The training procedure and different training choices are explained in section IV. Section V presents significant results of the proposed model in section II. Finally, section VI concludes the paper, providing some insights into future developments.

\newpage
\section{System Overview} \label{section:architecture}
In this section, the CNN architecture used to perform trajectories prediction is presented. 

\begin{figure}[ht!]
    \centering
    {
    \vspace{2mm}
    \includegraphics[width=0.97\columnwidth]{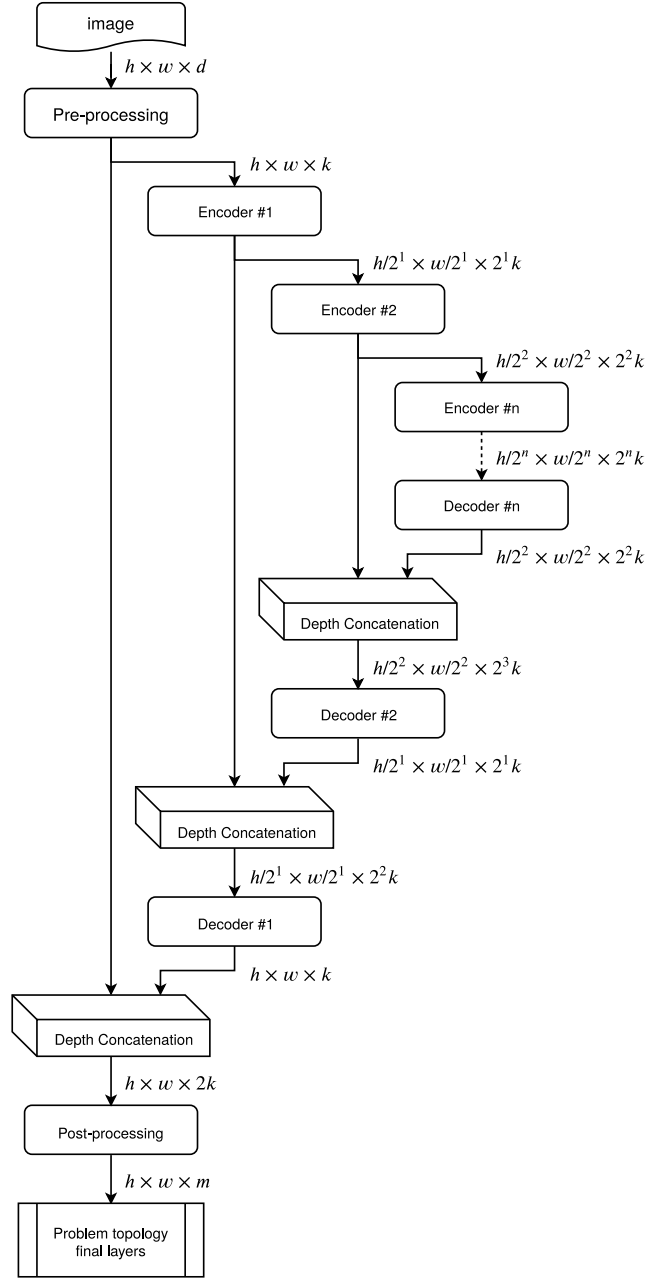}
    }
    \caption{U-net architecture based on deep levels.}
    \label{fig:unet}
\end{figure}

The U-net model \cite{unet} has been selected as the prediction core. This network was developed to perform semantic segmentation in biomedical imagery. This network presents different levels of features extracted from full image resolution to lower resolutions. All the features are combined in the end, adapting their sizes. The U-Net network is defined by the mainstream blocks that halves the original input image consecutively, then the same number of blocks doubles the size of the features block. The features extracted in the input side are concatenated with the features extracted in the output side at the corresponding levels. The number of levels is denoted as depth $n$. Fig. \ref{fig:unet} shows a simplified representation of the U-net's architecture. The pre-processing block generates $k$ features directly from the original image that will be double after each deep level.

In this approach, we use this model to perform image to image regression. The scene is represented into a Bird's Eye View (BEV) with dimensions $h\times w$. In the input side, $d$ representations of past samples are stacked, creating an image with $d$ channels. At the output side, an image with $d$ channels representing future samples is used as the network target. The idea is that the network core learns the underlying behavior presented in the input block and generates the same representation of that vehicles in a particular future point.

It is important to note that this network takes as input a data block with size at least $2^n \times 2^n$ and multiples of this value. Another critical point is to define the receptive filed from an output pixel. Eq. \ref{eq:inputreceptive} shows the influence range of a single input pixel where $n$ is the number of encoder-decoder blocks. Each encoder block expands the receptive field with two $3\times 3$ convolution layers, and them virtually multiplies by 2 with a \textit{maxPool} layer with a kernel size $2\times 2$. A decoder block has a \textit{2dtranspose} layer that increases the features by 2 in the two first dimensions. This block doubles the receptive field, then two convolution layers with kernel size $3 \times 3$ increase the contact surface.

\begin{equation}
\label{eq:inputreceptive}
r = \pm2 \left( 3+\sum_{i=2}^{n}{5\cdot 2^{i-2}}\right)
\end{equation}

For example, a U-net network with five deep levels would create a contact surface that links a pixel in the output layer on the position (0,0) with all the pixels in the input layer located at coordinate closer than (156,156). It would connect the output reference pixel to all pixels closer than (316, 316) for six levels of deep. Note that the output reference pixel is in the top left corner. The same connections are created to the left, right, up, and down. This point is critical due to the nature of vehicle interaction, especially in highway scenarios. Driver decisions, and consequently, vehicle trajectories are based on what the driver can see. In highway scenarios, the influence area of a certain vehicle grows according to its speed. The proposed architecture cannot ensure that the vehicle's computed position takes into account all that the driver is seeing. However, it takes into account up to 284 meters in both the front and backward directions, which covers all the study area when the vehicles are in the middle of the scene and the 75\% when the vehicle has just entered the study area. Table \ref{table:unetstruct} summarize main features of U-net model for common deep levels.

\begin{table}[!h]
\renewcommand{\arraystretch}{1.1}
    \centering
    \caption{U-net Contact Surface Area}
    \label{table:unetstruct}
    \begin{tabular}{ l | c | c | c | c |}
    Deep levels         & 4         & 5         & 6         & 7                 \\
    \hline
    Contact Area       & $\pm76$   & $\pm156$  & $\pm316$  & $\pm636$          \\
    Minimum input size  & 16        & 32        & 64        & 128               \\
    Parameters          & 56k       & 116k      & 235k      & 472k              \\
    \end{tabular}
\end{table}

\section{Data Transformation \label{section:io}}
In this section, the procedure followed to transform the HighD data into images is described in the first subsection. The reverse procedure to compute positions from images is detailed in subsection \ref{subsec:extractionalgo}. Finally, the association of generated positions to provided positions is specified in subsection\ref{subsec:association}.

\subsection{Input \& Output Representation}
The vehicle detections are represented in a BEV grid, where each pixel represents the probability of being a vehicle. Each vehicle is represented using a bi-dimensional Gaussian distribution, independently of the vehicle type. According to eq, the Gaussian distribution represents the probability of using a specific tile in the BEV. \ref{eq:gaussian}. The mean value of each bi-dimensional Gaussian distribution $\mathbf{\mu}$ is set using the center of the rectangle proposed for each detection. The standard deviation value $\mathbf{\sigma}$ is composed with the half of the width and height of the detection rectangle. When two vehicles overlap, the cumulative probability could overflow the maximum probability allowed. There are two options to merge the area shared by two vehicles. The first option is to add both and truncate it to the maximum probability. The other one is to use the maximum of both probability distribution functions according to eq. \ref{eq:probmax}. The second option represents the real scene in a more reliable way, and it is better for the position extraction algorithm as it will be exposed in subsection \ref{subsec:extractionalgo}.

\begin{equation}
p_i(x,y) = \exp{-\left(\left(\frac{x-\mu_{x_i}}{\sqrt{2}\sigma_{x_i}}\right)^2+\left(\frac{y-\mu_{y_i}}{\sqrt{2}\sigma_{y_i}}\right)^2\right)}
\label{eq:gaussian}
\end{equation}

\begin{equation}
p(x,y) = \max{\{ p_i(x,y)\}} \,\forall i
\label{eq:probmax}
\end{equation}

\begin{figure}[t]
    \centering
    {
    \vspace{2mm}
    \includegraphics[width=\columnwidth]{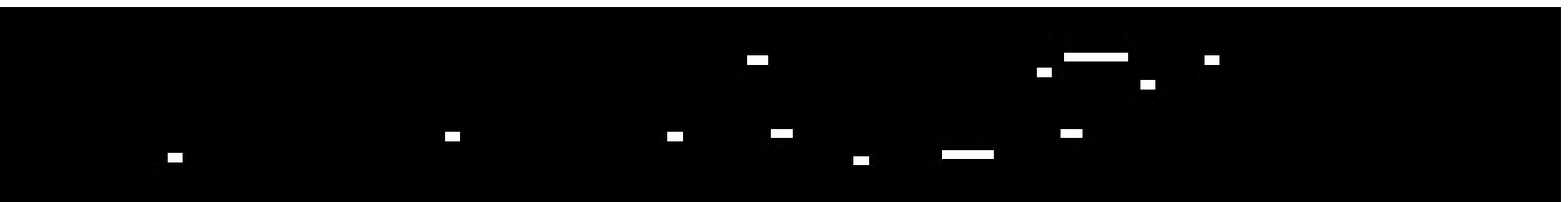}
    \includegraphics[width=\columnwidth]{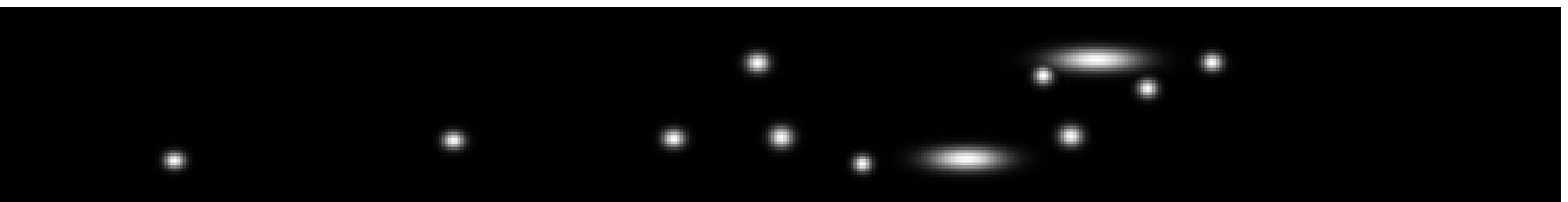}
    }
    \caption{Data representation in a BEV using rectangles (top) and Gaussian distributions (bottom). This representation corresponds to sequence 1, frame 1 of publicly available HighD dataset.}
    \label{fig:representation_example}
\end{figure}

Fig. \ref{fig:representation_example} illustrate a sample of the dataset. The top image represents the provided detections represented as rectangles using the original coordinates and dimensions. The bottom image represents the same scene but representing the vehicle detection making use of the Gaussian distribution function. Different vehicles can be observed with different lengths, corresponding to vehicles and trucks. This image represents simultaneously two traffic streams, one flowing from right to left in the upper part of the image and other flowing from left to right in the bottom.

The BEV representation has been defined as a grid with 512 pixels width and 64 pixels high. This grid represents a physical space of 512 meters along the $x$ axis and 32 meters along the $y$ axis according to the definition of the HighD coordinates. The grid's size has been established based on three criteria: memory size when it is allocated in the GPU for training purposes, a proper resolution to understand the scene, and compatibility with the proposed network architecture in \ref{section:architecture}.
The $y$ axis resolution is double than the $x$ resolution. This decision was taken because the longitudinal movement of vehicles is more stable and constant with more significant variations than lateral movement. A higher resolution in the $y$ axis allows detecting lateral behaviors easier.

Both input and output data are represented in the same way making no distinctions. The complete input and output data consist of $d$ consecutive samples stacked, creating an input/output volume with size $64 \times 512 \times d$.

\begin{figure}[h]
    \centering
    {
    \vspace{2mm}
    \includegraphics[width=\columnwidth]{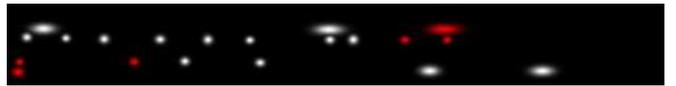}
    }
    \caption{Representation three seconds ahead from the last input representation. Red Gaussian distribution represents new and unexpected vehicles. Upper stream flows from right to left and lower from left to right. New vehicles are focus on both entries to the study area.}
    \label{fig:unexcpectedvehicles}
\end{figure}

When $d$ time samples of data are stacked, a new problem is generated in the output block. The output block represents future samples. Consequently, two kinds of vehicles coexist, ones who were present in the input block and others, which are new vehicles. The future positions of these new vehicles cannot be predicted as far as there are no data in the input block to consider their existence. Future scene representations contain only vehicles that were present in the last input representation. Fig. \ref{fig:unexcpectedvehicles} shows a scene representation three seconds ahead on time. White Gaussian distributions represent vehicles that were present three seconds back, and red ones represent new and not predictable vehicles.

\subsection{Vehicle Position Extraction}\label{subsec:extractionalgo}
The codification procedure transforms numeric data into a representation to make predictions. Once predictions are made, the representation needs to be transformed into numeric data. For each input image, $n$ different predictions are generated at different horizon time. The number of vehicles present in a future scene is \textit{a prioiri} unknown, so the way used to extract numeric positions of vehicles from each output representation must be able to produce a not fixed amount of positions. It can only consider the vehicles in the future should be the same or fewer than in the last known sample. 

The algorithm proposed in \ref{alg:positionextracion} is used to extract the position of the vehicles. The output data represents the probability of being using a tile in a certain future sample. The algorithm finds the pixel with the highest probability first. This pixel is denoted as $P=(R, C)$, and it is used as the discrete location of the vehicle. The discretization procedure used to represent both input and output data into a grid needs to be reverted. The proposed algorithm extracts the position with sub-pixel resolution in a second step. The position of the vehicle is refined using a scoring function. Each pixel $p_i$ included in the area defined by a rectangle with dimension $R = 2w \times 2h$ around $P$ contribute weighting its probability by its pixel coordinates according to eq. \ref{eq:scoringfcn}. Note that discrete positions are conditioned by the resolution used to define the probability occupancy map, which is set to 1.0 and 0.5 meters per pixel along the $x$ and $y$ axes.

\begin{equation}\label{eq:scoringfcn}
    \hat P(\hat{R},\hat{C}) = \sum_{r = R-h}^{r = R+h}{\sum_{c = C-w}^{c = C+w}{p(r,c)\cdot(r,c) }}\\
\end{equation}

\begin{figure}[ht!]
    \centering
    {
    \vspace{2mm}
    \includegraphics[width=\columnwidth]{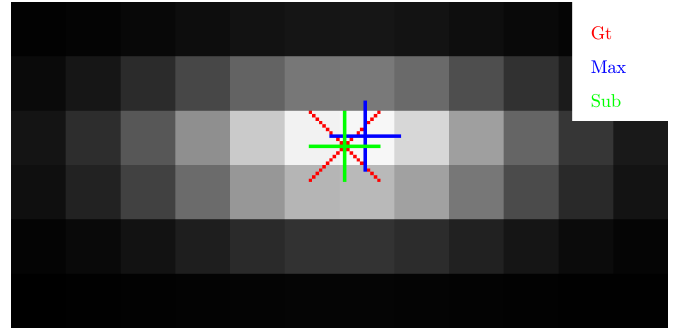}
    }
    \caption{Position extraction from graphic representation. The red cross represents the actual center of the vehicle, the blue plus symbol the discrete find position and the green plus symbol the sub-pixel position. Image augmented 16 times.}
    \label{fig:subpixel}
\end{figure}

After computing the sub-pixel position, the area used to compute it is cleared, setting the probability in the occupancy map to zero. This procedure is repeated as many times as tiles with a probability higher than $p_min$ remain in the occupancy map. According to the definition of the occupancy map, where a value of 1 means that the pixel is occupied and 0 means empty, we set the threshold $p_{min}$ to 0.5. That is the limit to consider that a pixel represents a possible vehicle. Fig. \ref{fig:subpixel} shows the codification of an arbitrary vehicle, the red cross represents the actual center of the vehicle, the blue plus symbol represents the discrete find position of the vehicle, and the green one represents the sub-pixel position of the vehicle. Note that the image has been zoomed by 16 to illustrate the differences between discrete and sub-pixel detection. Vehicle parameters are: $w = 5.0$, $h = 2.0$, $x = 6.63$, $y = 3.21$, and representation parameters: $x_{ppm}=y_{ppm}= 1$.

\begin{algorithm}
   \caption{Multi-Vehicle Location Extraction}\label{alg:positionextracion}
    \begin{algorithmic}[1]
      \Function{$\mathcal P =$ Extraction}{$p(r,c),p_{min},w,h$}
        \State $\mathcal{P} = \emptyset$
        \While{$\exists(r,c) \mid p(r,c) > p_{min}$}
            \State $P = (R,C) | p(R,C) > p(r,c) \forall(r,c)$
            \State $\hat{P}(\hat R ,\hat C) = SubpixelLocation(p(r,c), w, h, P)$
            \State $\hat{P} \in \mathcal{P}$
            \State $\forall(r,c)\in P \pm (h,w)$ do $p(r,c) = 0 $
        \EndWhile
      \EndFunction
      
      \Function{$ \hat P =$ SubpixelLocation}{$p(r,c), w, h, P$}
      \State $\hat R = \hat C = 0$
      \For{$R-h < r < R + h$} 
        \For{$C-w < c < C + w$}
            \State $\hat R += p(r,c)\cdot r$
            \State $\hat C += p(r,c)\cdot c$
        \EndFor
      \EndFor
      \State $\hat P=( \hat R / 2h, \hat C / 2w)$
      \EndFunction
\end{algorithmic}
\end{algorithm}

Table \ref{table:extraction} shows the position extracted from vehicle show in fig. \ref{fig:subpixel}. The vehicle is represented in a BEV using the same resolution in both axes, equal to 1 meter per pixel. The position of the vehicle is at coordinates $x,y=(6.63,3.21)$. The resolution used to make the representation defines the error generated when the maximum method is applied to extract the vehicle's position. However, the subpixel resolution method has errors in the order of tens of millimeters, and it is not related to the resolution used to represent the data.

\begin{table}[ht]
\renewcommand{\arraystretch}{1.1}
    \centering
    \caption{Position Extraction Methods}
    \label{table:extraction}
    \begin{tabular}{ l | c | c | c |}
                                & Original      & Maximum       & SubPixel                  \\
        \hline
        X / Y [m]/[m]           & 6.63 / 3.21   & 7 / 3         & 6.615 / 3.216             \\
        X / Y Error [m]/[m]     & - / -         & 0.37 / 0.21   & \textbf{0.015 / 0.006}    \\
    \end{tabular}
\end{table} 

\subsection{Association of Extracted Positions}\label{subsec:association}
\begin{figure}[t]
    \centering
    {
    \vspace{2mm}
    \includegraphics[width=\columnwidth]{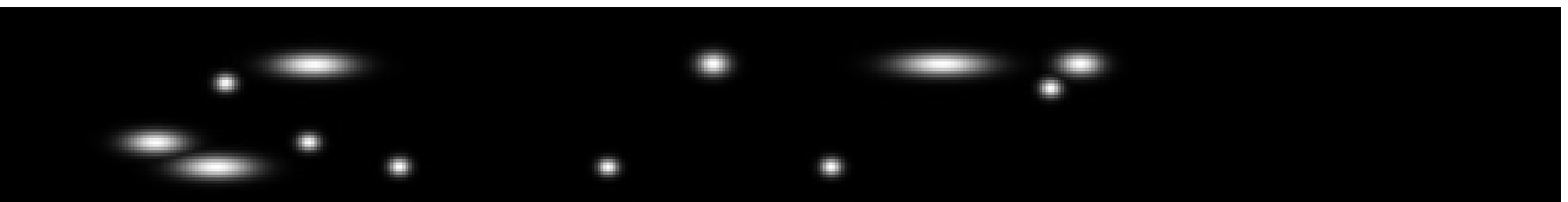}
    \includegraphics[width=\columnwidth]{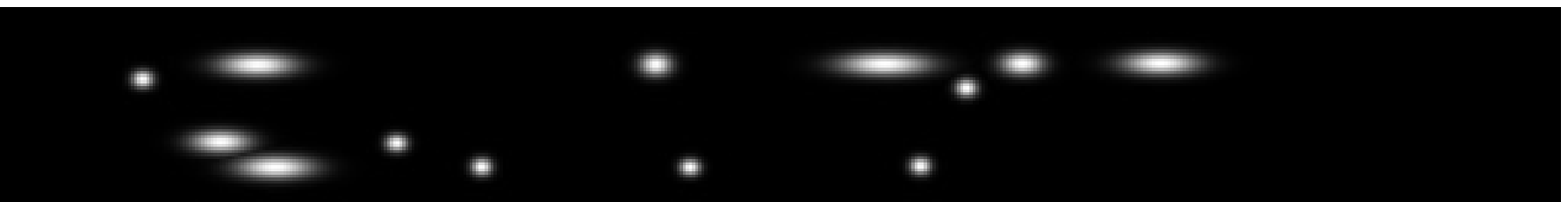}
    \includegraphics[width=\columnwidth]{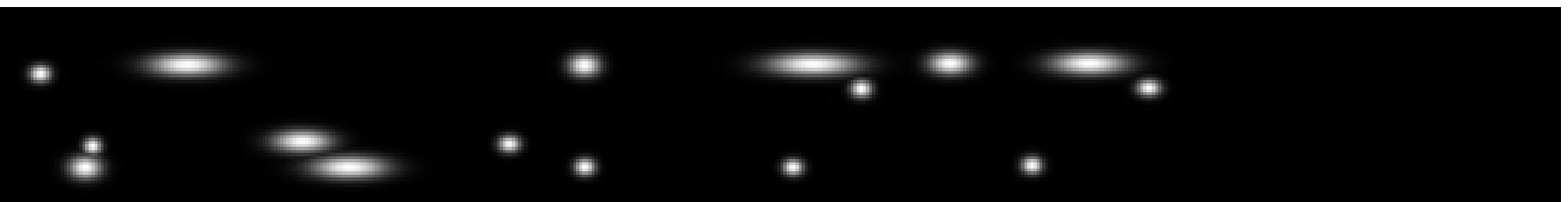}
    \includegraphics[width=\columnwidth]{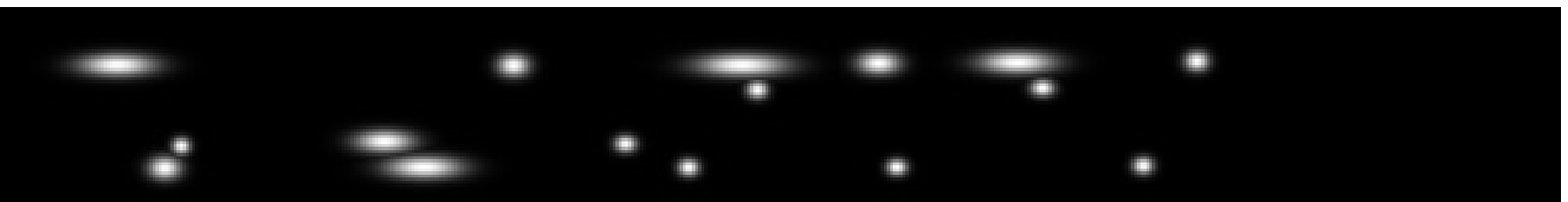}
    \includegraphics[width=\columnwidth]{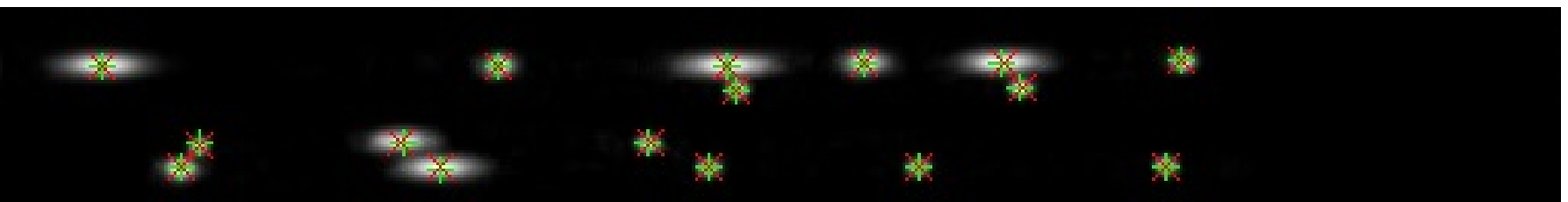}
    \includegraphics[width=\columnwidth]{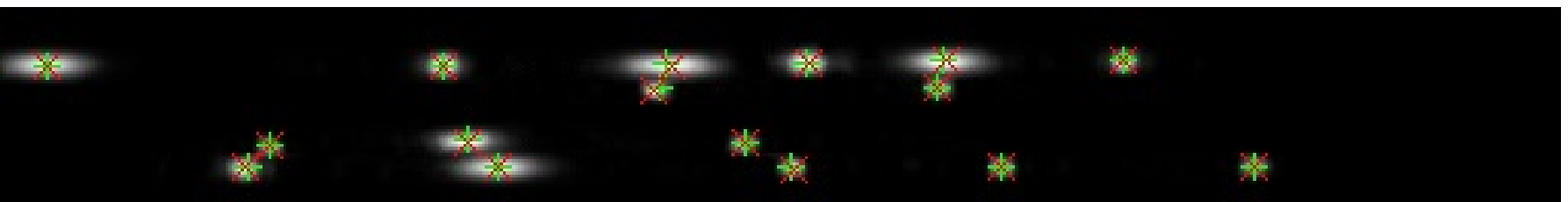}
    \includegraphics[width=\columnwidth]{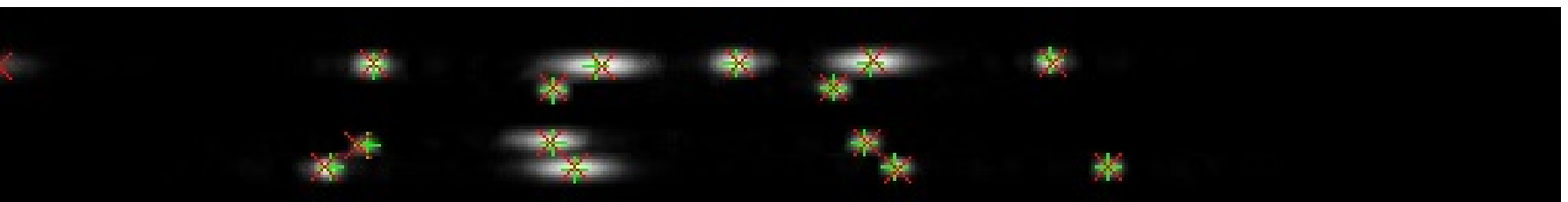}
    \includegraphics[width=\columnwidth]{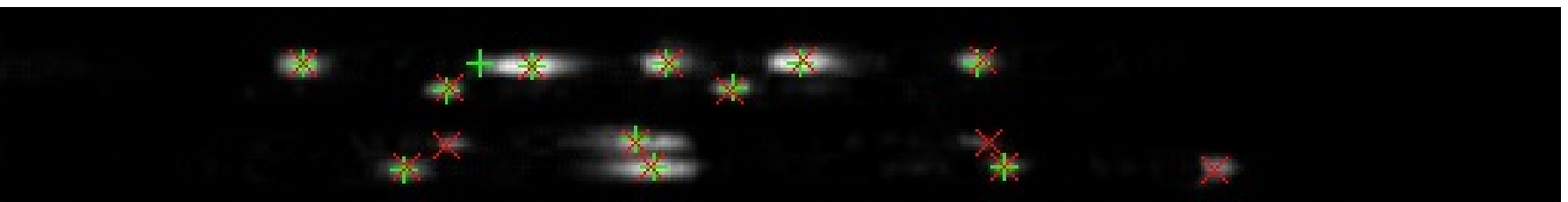}
    }
    \caption{Data representation in a BEV of predicted vehicles. Four top images represent a input sequence data, four bottom represents predictions at $t = \{0.2,1.0,2.0,3.0\}$ seconds ahead. Red crosses represent the position of the prediction targets, and green plus symbols are the predicted positions extracted from the image generated by the U-net model.}
    \label{fig:final}
\end{figure}

When the positions of all the vehicles are extracted from an image, they need to be associated with their respective detection. A simple procedure based on a Hungarian matrix \cite{kuhn1955kuhn} is used to associate the extracted positions with the positions given in the dataset. The number of elements that can be matched is the minimum between the number of positions extracted from the image of the number of detections provided for the corresponding scene. The value used as the distance parameter to associate elements is the euclidean distance between extracted points and detections. This method is good enough as predicted positions do not differ from the actual positions of the prediction targets. Fig. \ref{fig:final} shows predicted positions and associate target prediction. As can be seen, the association process works well for this type of problem. An element in the last prediction image does not make with a target because it has not be predicted due to a low probability of existence.

\section{Training Strategy \label{section:training}}
The available data in the HighD dataset consists of 60 sequences with lengths starting from 9700 up to 32200 frames recorded at 25 Hz. The number of vehicles present in these sequences accounts for 110K in a total of 1500K frames representing more than 16 hours of traffic recordings. The amount of data is massive and allows us to train models with various samples and situations. The frame rate is too high to appreciate differences from one frame to the next with the resolution used to represent the scene. The original frame rate was reduced from 25Hz to 5Hz, removing four of each five samples. The input data stack 15 BEV representations of past samples. The output stack 15 BEV representations of the next consecutive samples. The lowered time resolution allows the input and output data to cover bigger spam of time using the same number of channels. The time represented in the input block is from $t$ to $t - 2.8$ seconds. The output block represents future locations from $t+0.2$ to $t + 3.0$ seconds.

\subsection{Training Hyper-Parameters}
The training set contains only the samples included in sequences from 1 to 20  due to the vast amount of data. Samples from sequences 21 to 25 have been used as a test set. Table \ref{table:datasumary} summarize the main features of the dataset, training, and test set used to develop the models.

\begin{table}[ht]
\renewcommand{\arraystretch}{1.1}
    \centering
    \caption{Dataset, Training, \& Test Stats}
    \label{table:datasumary}
    \begin{tabular}{ l | c | c | c |}
                                & Samples @ 25Hz    & Samples @ 5Hz  & Trajectories \\
        \hline
        HighD dataset           & 1500K             & 300K           & 110K         \\
        Training set            & 463K              & 93K            & 28K          \\
        Test set                & 126K              & 25K            & 7K          \\
    \end{tabular}
\end{table} 

The influence of high-level parameters such as the network's deep level and the output topology has been tested, doing different trainings varying them. The results of each variation is presented in section \ref{sec:results} in table \ref{table:results_unet}.
Regarding the deep levels of the U-net, it was limited to 5 and 6. Levels below five cannot be tested because of input size restrictions, as specified in table \ref{table:unetstruct}. Level 7 and higher exceeds the GPU memory size, and the training could not be conducted.
The last layers of the U-net were replaced to fit the regression problem. Three different output layers were identified as possible output layers: \textit{linerLayer}, \textit{tanh}, \textit{clippedReLu}. The linear layer does not apply a transformation to the network's output. The \textit{tanh} layer applies the hyperbolic tangent function ranging the output into the range of $(-1,1)$ with a non-linear transformation. The \textit{clippedRelu} keeps the output between 0 and a given value, which is 1 for this purpose. The last one seems to be perfect to fit the output problem with its values in the range $(0,1)$.

The main training hyper-parameters are detailed here: mini-batch size = 1, epoch = 1, initial learning rate $10^{-6}$, momentum = 0.9, gradient threshold = 1, loss function = MSE. 

\section{Results}\label{sec:results}
This section presents and discusses the final results generated by the proposed model described in sections \ref{section:architecture} and \ref{section:io} to predict the trajectories of vehicles in a group way. For a better understanding of training times and inference rates, the details of the computer used to carry out these experiments are given. PC with Kubuntu 18.04LTS, i7-7700K CPU, 32GB of RAM, and NVIDIA GTX1080Ti GPU using Matlab 2019b.

For trajectory prediction problem evaluation, we used the two metrics to evaluate the goodness of each configuration. Longitudinal and lateral absolute errors are computed for each prediction time. Table \ref{table:results_unet} shows the mean values of these errors at two points: the first prediction sample and the last one, equivalent 0.2 and 3 seconds. For clarity, we omitted other prediction intervals. The configuration with the \textit{tanh} layer produces output images where any vehicle position could be extracted.  
\begin{table}[h]
\renewcommand{\arraystretch}{1.1}
    \centering
    \caption{Prediction Error by Network Topology}
    \label{table:results_unet}
    \begin{tabular}{ l | c | c | }
                                        & $t = 0.2$                                  & $t = 3.0$  \\
        U-net topology                  & $\varepsilon_x/\varepsilon_y$    & $\varepsilon_x/\varepsilon_y$  \\
        \hline
        Deep = 5, fcn = linear          & 0.52 / 0.17               & 2.36 / 0.54           \\
        \textbf{Deep = 6, fcn = linear}   & \textbf{0.23 / 0.01} & \textbf{1.23 / 0.07}           \\
        Deep = 5, fcn = tanh            & - / -                     & - / -                 \\
        Deep = 6, fcn = tanh            & - / -                     & - / -                 \\
        Deep = 5, fcn = clippedRelu     & 0.74 / 0.38               & 2.51 / 0.94           \\
        Deep = 6, fcn = clippedRelu     & 0.46 / 0.22               & 2.06 / 0.62           \\
    \end{tabular}
\end{table}

\begin{table}[h]
\renewcommand{\arraystretch}{1.1}
    \centering
    \caption{Prediction Error by Network Topology}
    \label{table:results_unet}
    \begin{tabular}{ l | c | c | }
                                        & $t = 0.25$                                  & $t = 2.0$  \\
        U-net topology                  & $\varepsilon_x/\varepsilon_y$    & $\varepsilon_x/\varepsilon_y$  \\
        \hline
        Deep = 5, fcn = linear          & 0.65 / 0.21               & 1.57 / 0.36           \\
        \textbf{Deep = 6, fcn = linear}   & \textbf{0.29 / 0.01} & \textbf{0.82 / 0.04}           \\
        Deep = 5, fcn = tanh            & - / -                     & - / -                 \\
        Deep = 6, fcn = tanh            & - / -                     & - / -                 \\
        Deep = 5, fcn = clippedRelu     & 0.92 / 0.47               & 1.67 / 0.63           \\
        Deep = 6, fcn = clippedRelu     & 0.57 / 0.27               & 1.37 / 0.41           \\
    \end{tabular}
\end{table} 

The training evolution reduced the prediction error quickly and progressively until the last iteration in the best-case-scenario. This progressive reduction indicates that the model is generalizing because new and not view by the net samples produces lower errors than previous ones. The training time took around 4 and 23 hours for the 5 and 6 deep levels, respectively. The different output layers do not produce a significant delay in the training time. Both models can run at 63 and 16 Hz at deploy time, respectively, accounting only the inference time.    


\section{Conclusions and Future Work}
As conclusions, a novel method to predict trajectories in a group way, including all the involved elements in a simple BEV representation is presented. Motion histories and vehicle interactions are included by stacking consecutive representations of past samples. The U-net is used as a prediction core, which has shown an impressive capacity to generate visual representations of future positions. The use of Gaussian distributions to represent the vehicles allows us to retrieve subpixel positions from discrete images.

The network at it is only can be used to predict positions 3 seconds ahead. However, the proposed architecture has the same dimensions in the input and the output sides. The first channel of the output block can be used as the current sample in the input block and the new output extends the prediction one step more. This process can be repeated as many times as desired, but the output quality becomes lower due to the degradation of the input data. The inference time of this model does not vary with the number of involved vehicles.

It has been observed that the U-net model with five deep levels does not perform good quality outputs; the authors consider that it is related to the receptive field created by the network. It has been observed that large vehicles such as trucks produce double detections when the position extraction method is used. It is caused because the fixed-sized used to clear the occupancy map does not clear all the areas occupied by the truck. This problem can be solved by using the ground truth's information to set the width and height of the vehicles, but it cannot be used at deploy time.

As future work, the efforts must be focused on increase the training data and extend the training time. The vast amount of data present in the HighD and the number of conducted tests took many days. The two streams of traffic can be used to double the samples generating a unique traffic flow direction. U-net with seven deep levels seems promising, a by-parts training can be conducted to train models that do not fit in the GPU memory. Some trainings were conducted using a mini-batch size different to 1. The output images have less definition than the same trainings with a single image mini-batch.

\newpage
\section*{ACKNOWLEDGMENT}
This work was funded by Research Grants S2018/EMT-4362  (Community Reg. Madrid), DPI2017-90035-R  (Spanish Min. of Science and Innovation), BRAVE Project, H2020, Contract \#723021. It has also received funding from the Electronic Component Systems for European Leadership Joint Undertaking under grant agreement No 737469 (AutoDrive Project). This Joint Undertaking receives support from the European Unions Horizon 2020 research and innovation programme and Germany, Austria, Spain, Italy, Latvia, Belgium, Netherlands, Sweden, Finland, Lithuania, Czech Republic, Romania, Norway, and FPU14/02694 (Spanish Min. of Education) via a predoctoral grant to the first author.

\bibliographystyle{bibliography/IEEEtran}
\bibliography{bibliography/IEEEabrv,bibliography/references}

\end{document}